\documentclass[conference, compsoc, twoside]{IEEEtran}
\usepackage{fancyhdr}
\usepackage[utf8]{inputenc}
\usepackage{graphicx}
\usepackage{csquotes}
\usepackage{enumitem}
\usepackage{cleveref}
\usepackage{amsmath}
\usepackage{diagbox}
\usepackage{array}
\usepackage{hhline}
\usepackage{xcolor}
\usepackage{colortbl}
\usepackage{multirow}
\usepackage{multicol, blindtext}

\newcommand{\etc}{\emph{etc.}}

\ifCLASSOPTIONcompsoc
  \usepackage[nocompress]{cite}
\else
  \usepackage{cite}
\fi

\ifCLASSINFOpdf
\else
\fi

\begin{document}

\title{Learning spectro-temporal features \\with 3D CNNs for speech emotion recognition}

\author{\IEEEauthorblockN{Jaebok Kim, Khiet P. Truong, Gwenn Englebienne, and Vanessa Evers}\\

\IEEEauthorblockA{Human Media Interaction, University of Twente, Enschede, The Netherlands\\
Email: \{j.kim, k.p.truong, g.englebienne, v.evers\}@utwente.nl}
}

\maketitle
\thispagestyle{fancy}
\begin{abstract}
In this paper, we propose to use deep 3-dimensional convolutional networks (3D CNNs) in order to address the challenge of modelling spectro-temporal dynamics for speech emotion recognition (SER). Compared to a hybrid of Convolutional Neural Network and Long-Short-Term-Memory (CNN-LSTM), our proposed 3D CNNs simultaneously extract short-term and long-term spectral features with a moderate number of parameters. We evaluated our proposed and other state-of-the-art methods in a speaker-independent manner using aggregated corpora that give a large and diverse set of speakers. We found that 1) shallow temporal and moderately deep spectral kernels of a homogeneous architecture are optimal for the task; and 2) our 3D CNNs are more effective for spectro-temporal feature learning compared to other methods. Finally, we visualised the feature space obtained with our proposed method using t-distributed stochastic neighbour embedding (T-SNE) and could observe distinct clusters of emotions.
\end{abstract}


%
\IEEEpeerreviewmaketitle

\section{Introduction}

Recently, deep learning methods such as Fully-connected Neural Networks (FCN)~\cite{kunHan2014dnn}, Convolutional Neural Networks (CNN)~\cite{zheng2015experimental,mao2014learning}, and Long Short-Term Memory (LSTM)~\cite{trigeorgis2016adieu,lee2015high} have shown considerable improvements of performance in speech emotion recognition (SER). As a potential way to improve performance, representation learning has been used to build high-level features from low-level features through several layers \cite{bengio2013representation,zheng2015experimental,mao2014learning}. However, learning sequential structures of spectrogram representations appeared to be still challenging \cite{ghoshrepresentation2016}. CNN-based methods have been investigated to this end~\cite{zheng2015experimental,mao2014learning,anandconvoluted,trigeorgis2016adieu}. Although a hybrid of CNN and LSTM can be a promising method to deal with spectral variations and temporal dependencies, LSTM has the limitation of increasing depth caused by the great number of parameters. Hence, learning temporal dynamics of spectral properties for SER remains a challenge.

In this paper, we propose to learn spectro-temporal features using deep 3D CNNs. 3D CNNs are able to extract spatio-temporal features in a seamless way and have shown promising performances in computer vision tasks~\cite{ji20133d,tran2015learning}. For the task of SER, our proposed method composes a temporal series of 2D spectral feature maps and models both short and long-term dependencies simultaneously. Using large-scale datasets (7 representative, aggregated corpora) and speaker-independent classification experiments, we evaluated our proposed 3D CNNs for SER in a way that is representative of challenges for SER ``in the wild''. We found that 1) homogeneous layers with shallow temporal but deep spectral kernels work best among the limited set of explored architectures, and that 2) our proposed 3D CNNs are more effective and efficient for spectro-temporal feature learning in SER compared to other CNN-based methods.

This paper is structured as follows. We first introduce related studies in Section \ref{sec:relatedwork}. Next, we present corpora in Section~\ref{sec:data}, and describe our proposed learning method in Section \ref{sec:method}. The results will be reported in Section~\ref{sec:result} and concluded in Section~\ref{sec:conclusion}.

\section{Related Work}\label{sec:relatedwork}

The performance of machine learning relies on ``feature engineering'' that puts large effort in finding an optimal set of features. For the same reason, previous works in SER has focused on finding optimal feature sets and resulted in the wide usage of off-the-shelf features such as F0, Mel-Frequency Cepstrum Coefficients, and energy. However, the performance greatly varies between corpora that have distinct tasks. Representation learning offers a partial and potential solution by extracting high-level features from low-level features through a composition of multiple non-linear transformations~\cite{bengio2013representation}. With deep architectures, representation learning offers two advantages: 1) re-use of features which yields benefits in both computational and statistical efficiency, and 2) abstraction of features. For example, CNN can extract abstract features in a more explicit way via a pooling mechanism \cite{lecun1995convolutional}. 

The performance of SER using deep architectures can still be much improved, and an optimal feature set has not been found yet for SER. For example, in~\cite{kunHan2014dnn,lee2015high,kim2017interspeech}, high-level features obtained from off-the-shelf features outperformed conventional methods. However, representation learning using log-spectrogram features did not outperform that of using off-the-shelf features - learning such a complex sequential structure of emotional speech appeared to be hard for representation learning \cite{ghoshrepresentation2016}. 

More recently, the field of SER started employing CNN using low-level features. CNN is able to recognise patterns with distortions and variations \cite{lecun1995convolutional}. It operates two functions and generates the integral of the point-wise multiplication of the two functions. For one dimensional (D) discrete data, it is defined as:

\begin{equation} \label{eq:highway-1}
  (f \ast g)(t) = \displaystyle \sum_{k=-T}^{T} f(t-k) \cdot g(k)
\end{equation}

CNN-based methods using low-level features were proposed and outperformed off-the-shelf feature-based methods \cite{zheng2015experimental,mao2014learning,anandconvoluted,trigeorgis2016adieu,kim2017acmmm}. In~\cite{zheng2015experimental,mao2014learning,anandconvoluted} 2D feature maps were composed of spectrogram features with a fine resolution. However, these 2D CNNs cannot model temporal dependency directly. Instead, LSTM should be followed to model temporal dependencies \cite{anandconvoluted,trigeorgis2016adieu}. Moreover, temporal convolutions can extract spectral features from raw wave signals and capture long-term dependencies \cite{trigeorgis2016adieu}. Lastly, CNN-LSTM-DNN was proposed to address frequency variations in spectral domain, long-term dependencies, separation in utterance-level feature space for the task of speech recognition \cite{sainath2015convolutional}. While these methods augment CNNs and LSTM to handle spectral variations and temporal dynamics, a large number of parameters are required, and it is hard to learn complex dynamics with limited depths. Without these complex memory mechanisms, 3D CNNs could learn temporal features \cite{ji20133d,tran2015learning}. In ~\cite{ji20133d,tran2015learning}, a series of human's motion was modelled by 3D CNNs, it empirically turned out that 3D CNNs are not only effective but also efficient to capture spatio-temporal features.

\section{Data}\label{sec:data}

We select seven representative corpora: LDC Emotional Prosody~\cite{liberman2002emotional}, eNTERFACE~\cite{martin2006enterface}, EMODB~\cite{burkhardt2005database} FAU-aibo emotion corpus~\cite{batliner2004you}, IEMOCAP~\cite{busso2008iemocap}, SEMAINE~\cite{mckeown2010semaine}, and RECOLA~\cite{ringeval2013introducing}. Since there are more corpora that have discrete labels than those that have continuous labels (e.g. arousal and valence), we focus on four discrete categories, neutral, happy, sad, and angry, which are commonly accessible as summarised in Table \ref{tab:corpora}. However, the SEMAINE and RECOLA corpora provide only continuous labels such as arousal and valence, not discrete categories. To map the continuous labels into the four discrete categories, we use the landmarks of the valence and arousal dimensions as provided in FEELTRACE\cite{cowie2000feeltrace}. We extract segments by using voice activity detection or given time-alignment labels. Then, we calculate the Euclidean Distance between the landmarks and the values of the valence and arousal dimensions of each segment. Since each segment has a sequence of values of valance and arousal, we calculate the average distance for each discrete category. Next, we assign the emotional category with the smallest (average) distance to the valence and arousal values. Table \ref{tab:feeltrace} shows the Feeltrace landmarks for the four categories and the corresponding valence and arousal values. We use only speech utterances mapped into the four emotional categories and remove those with other categories. To the best of our knowledge, our dataset obtained from aggregating the 7 corpora has the largest number of speakers and samples in the deep-learning based experiments for SER.

\begin{table}[!t]
\centering
\begin{tabular}{lrrrrr}

\hline
Corpus ID & Speakers  & \multicolumn{4}{c}{Emotion} \\ \cline{3-6} 
  & & neutral & happy & sad & angry \\
\hline
AIBO    & 51 & 10967  & 889 & 0 & 1492  \\
EMODB     & 10 & 77 & 61  & 58 & 97     \\
ENTERFACE   & 43 & 0  & 208 & 422 & 211     \\
LDC     &   7 & 80  & 180 & 161 & 139     \\
IEMOCAP   &   10  & 1708  & 595 & 2168 & 2206   \\
SEMAINE   & 20  & 2694  & 766 & 82 & 392  \\
RECOLA    & 23 & 159 & 14 &  109 & 121 \\
\hline
Total   & 164 & 15685 & 2713  & 3000  & 4658  \\
\hline
\end{tabular}
\vspace{0.15cm}
\caption{Overview of the selected corpora (the number of speakers and utterances)}\label{tab:corpora}
\end{table}

\begin{table}[!t]

\centering
\vspace{10px}
\begin{tabular}{ccc}
\hline
discrete emotional categories   & valence & arousal\\
\hline
neutral   & 0.00 & 0.00 \\
happy   & 0.74  & 0.52\\
angry   & -0.77 & 0.75\\
sad     & -0.7  & -0.48\\
\hline
\end{tabular}
\vspace{0.15cm}
\caption{Landmarks for the SEMAINE and RECOLRA corpora in FEELTRACE.}\label{tab:feeltrace}
\end{table}

\section{Method}\label{sec:method}

\begin{figure}[!tb]
\begin{minipage}[b]{0.48\linewidth}
  \centerline{\includegraphics[width=3.8cm]{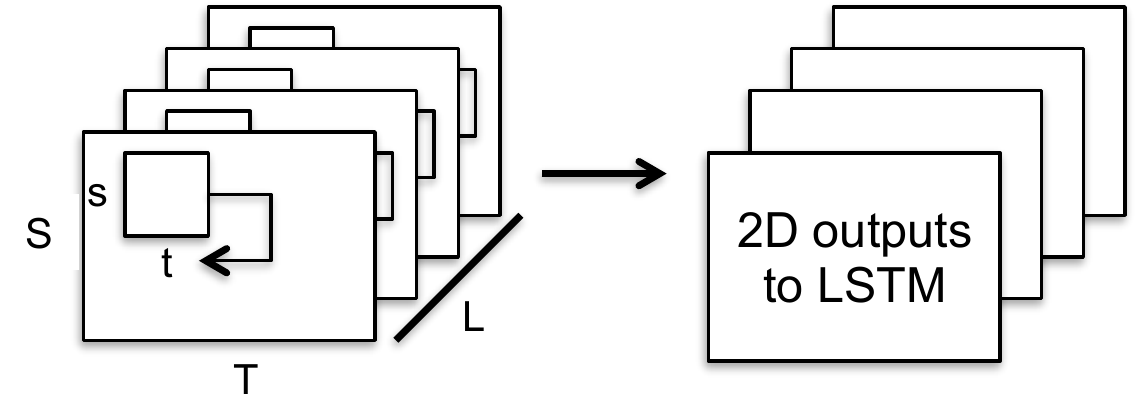}}
  \centering{(a) 2D-CNN-LSTM}
\end{minipage}
\begin{minipage}[b]{0.48\linewidth}
  \centerline{\includegraphics[width=3.8cm]{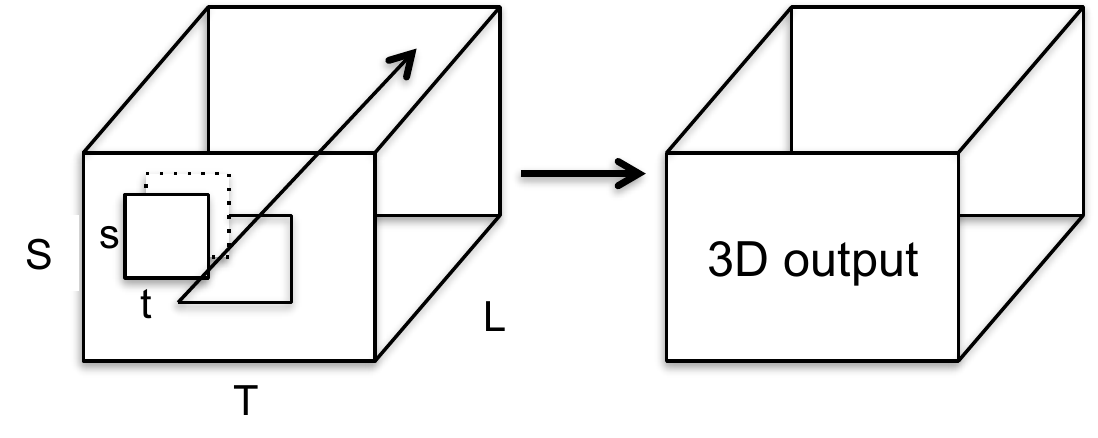}}
  \centering{(b) 3D-CNN}
\end{minipage}
\vspace{0.15cm}
\caption{The comparison of 2D-CNN-LSTM and 3D-CNN: a) Applying 2D convolution on each spectral feature map (T x S) in a time series (L). b) Applying 3D convolution on a time series of maps (L x T x S) results in another volume, preserving temporal dynamics of the input.}
\label{fig:cnn}
\end{figure}

While the previous methods~\cite{anandconvoluted,sainath2015convolutional} learn spectral features and the temporal dependencies via the augmentation of CNN and LSTM, our proposed method is designed to learn spectro-temporal features simultaneously as depicted in Figure \ref{fig:cnn}. Particularly, spectral features should have a sufficiently fine resolution, and both short-term ($\sim$ 200ms) and long-term ($\sim$ 2s) should be considered. To this end, the following network topology is configured.

\textbf{Input feature maps.} First, we segment utterances into 2s long sequences after min-max normalisation of gains per speaker. Zero-padding is applied for utterances shorter than 2s while we trim those longer than 2s. Then, we extract 256 points log-spectrogram every 20ms. Since it potentially causes over-fitting, we do not use a sliding contextual window on it. Therefore, we obtain 100 frames. Lastly, we compose a temporal series of 2D feature maps that have a resolution of 10 x 256. Each feature map represents spectral features in shot-term (200ms) windows. As a result, each utterance segment has a resolution of 10 x 10 x 256. Let us denote elements of the resolution as long-term (L), short-term (T), and spectral (S).

\textbf{3D CNNs.} Input feature maps are directly fed into 3D CNNs. Based on the previous finding~\cite{tran2015learning}, we adopt a homogeneous architecture, i.e., all convolutional layers have the same resolutions (L x T x S). In subsequent experiments, we empirically find the optimal resolutions and the number of convolutional layers for our task. All convolutional layers have four kernels.

\textbf{3D Max-pooling.} A 3D max-pooling layer follows each 3D CNN. However, to preserve the spectro-temporal features at early phases~\cite{tran2015learning}, we do not pool outputs of the first convolutional layers. Hence, except for the first pooling layer, the rest of the pooling layers have the resolution of 2 x 2 x 2. 

Lastly, we examine two methods to learn utterance-level spectro-temporal features: (a) we simply flatten 3D volume output features into 1D volume output vectors that are fed into fully-connected layers with a softmax layer; and (b) we transform 3D volume output features into 2D volume output features that are fed into a temporal series of fully-connected layers. Both methods use two fully-connected layers with 512 nodes. Let us denote method (a) and (b) as \textbf{3D-CNN-DNN} and \textbf{3D-CNN-DNN-ELM}.

\begin{figure}[!t]
\begin{minipage}[b]{1.0\linewidth}
  \centering
  \centerline{\includegraphics[height=5.0cm,width=7.8cm]{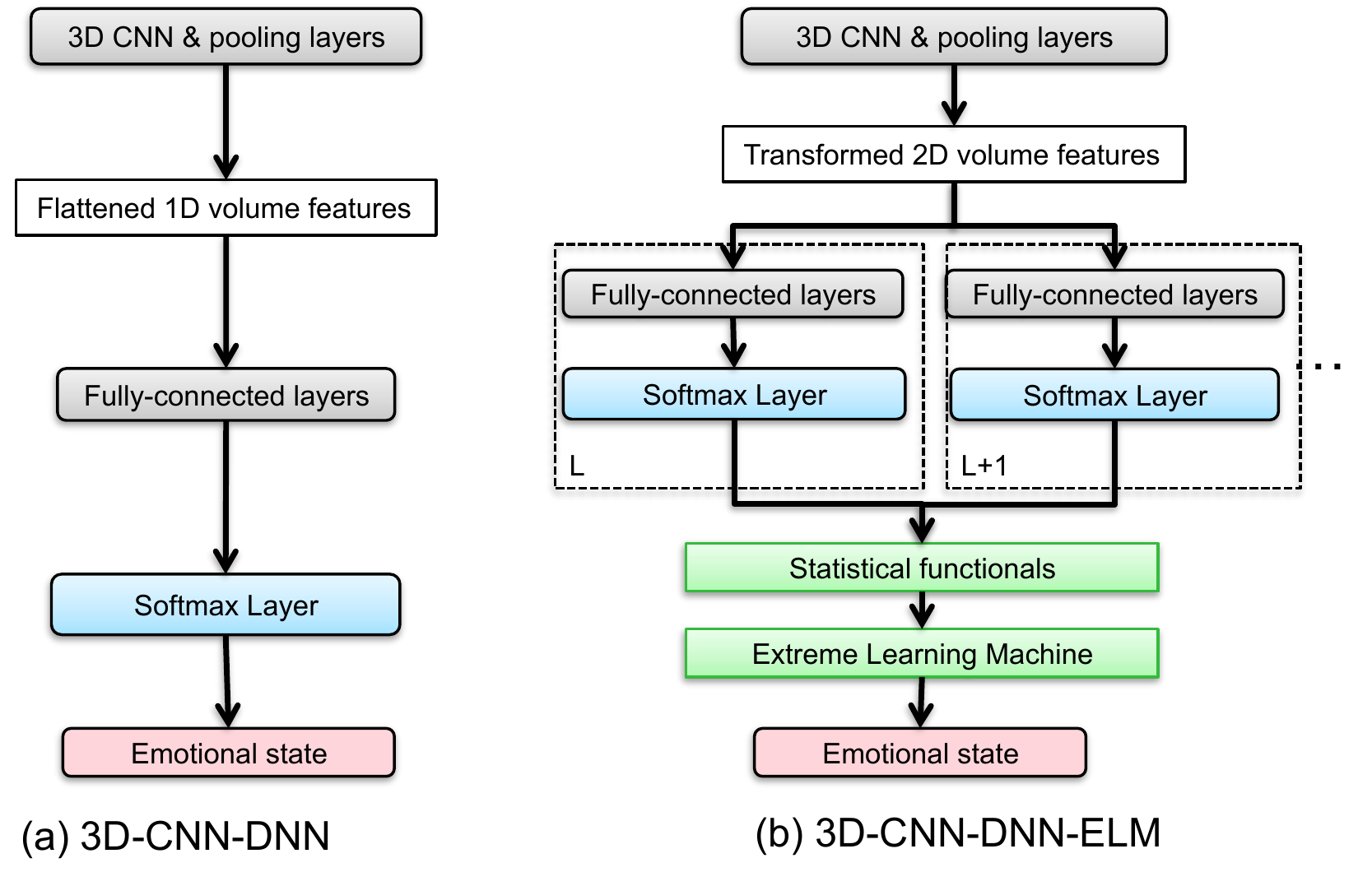}}
\end{minipage}
\caption{Block diagram of the proposed methods; L (state in long-term temporal resolution)}
\label{fig:topology}
\end{figure}

3D-CNN-DNN-ELM does not use pooling for the long-term depth to train a sequence of the fully-connected layers. Moreover, it requires statistical functionals at the softmax layer and a proceeding classifier such as Extreme Learning Machine (ELM) as proposed in \cite{kunHan2014dnn,lee2015high}. We follow the same set-up of the functionals and ELM proposed in \cite{kunHan2014dnn,lee2015high}. Training a linear classifier on features from the top fully-connected layer of 3D-CNN-DNN can be an effective approach~\cite{tran2015learning} but it may bring similar effects of 3D-CNN-DNN-ELM. Hence, we do not examine the variant of 3D-CNN-DNN. Figure \ref{fig:topology} illustrates the difference between 3D-CNN-DNN and 3D-CNN-DNN-ELM. 

\section{Experiments and Results}\label{sec:result}
\begin{table}[!th]
\centering
\begin{tabular}{llllll}
\hline
Shapes & Resolutions     & Best & Worst & Mean \\
\hline
(a) shallow temporal \& & 2 x 2 x 2          & $.518$   & $.451$ & $.480\pm.03$ \\
deep spectral kernels   & 2 x 2 x 32         & $.541$   & $.425$ & $.484\pm.05$ \\
                        & 2 x 2 x 128        & $.550$   & $.471$ & $\mathbf{.496\pm.03}$ \\    
                        & 2 x 2 x 256        & $.516$   & $.250$ & $.416\pm.10$ \\    
\hline
(b) deep temporal \&  & 4 x 2 x 2          & $.481$   & $.393$ & $.450\pm.04$ \\
shallow spectral kernels & 8 x 2 x 2          & $.507$   & $.394$ & $.454\pm.04$ \\
                             & 2 x 4 x 2          & $.503$   & $.419$ & $.450\pm.03$ \\
                             & 2 x 8 x 2          & $.487$   & $.441$ & $\mathbf{.463\pm.02}$ \\
\hline
(c) deep temporal \& & 4 x 2 x 128        & $.469$   & $.437$ & $.451\pm.01$ \\   
deep spectral kernels     & 8 x 2 x 128        & $.528$   & $.421$ & $.462\pm.04$ \\   
                     & 2 x 4 x 128        & $.492$   & $.446$ & $\mathbf{.467\pm.02}$ \\   
                     & 2 x 8 x 128        & $.488$   & $.423$ & $.467\pm.03$ \\
\hline
\end{tabular}
\vspace{0.15cm}
\caption{Best, worst, and mean performance (UA) by varying resolutions (L x S x T) of 3D kernels}\label{tab:resolution}
\end{table}

First, we investigate how resolutions of kernels affect performance of the 3D CNNs. Next, based on the optimal resolution, we compare our proposed method to other state-of-the-art methods using various features and architectures. As a common set-up, we use Adam method~\cite{kingma2014adam} with a mini-batch of 128 samples, and a fixed learning rate of $3 \cdot 10^{-3}$. We use categorical cross-entropy as the cost function. To prevent over-fitting, we use early-stopping \cite{prechelt1998automatic} (the maximum number of epoch: $20$) and dropout~\cite{srivastava2014dropout} ($p =.5$) for fully-connected layers. As an evaluation metric, Unweighted Accuracy (UA) is used to consider the imbalanced distribution of the classes. We aim to conduct cross-validation while keeping speaker-independence. Hence, we compose 5-fold cross-validation. First, we independently shuffle speakers in each corpus. Then, we divide each corpus into testing (20\%), validation (20\%), and training (60\% of the total number of the speakers) sets. Next, seven testing sets (from the seven corpora) are merged as testing data for one fold. Validation and training data for the fold are constructed in the same way. We repeat this process five times to build 5 folds. Lastly, we use Wilcoxon signed-rank paired test ($0.99$ of confidence level and two-sided tests)~\cite{gibbons2011nonparametric} to check significance of gains.

\begin{table*}[!th]
\centering
\begin{tabular}{lllll}
\hline
Features & Methods            &   Configuration of layers         & Parameters (K) & UA \\
\hline
off-the-shelf & DNN-ELM~\cite{kunHan2014dnn}      & 3 x FCN (256) + ELM   & $141$ & $.384\pm.05$\\
        & LSTM-ELM~\cite{lee2015high}       & 2 x LSTM (128) + ELM  & $214$ & $.421\pm.11$\\
\hline
raw waveform & 1D-CNN-LSTM~\cite{trigeorgis2016adieu}  & CNN (80) + CNN (800) + 2 x LSTM (128) & $1583$ & $.391\pm.06$\\
        & & CNN (80) + CNN (800) + 2 x LSTM (256) & $2211$ & $.380\pm.03$\\   
\hline
log-spectrogram & 2D-CNN-LSTM~\cite{anandconvoluted}  & 2 x CNN (2 x 2) + 2 x LSTM (128) & $264$ & $.318\pm.01$\\
        &         & 2 x CNN (2 x 32) + 2 x LSTM (128) & $268$ & $.323\pm01$\\
        &         & 2 x CNN (2 x 128) + 2 x LSTM (128) & $282$ & $.306\pm.03$\\ 
        \cline{2-5}
        & 2D-CNN-LSTM-DNN~\cite{sainath2015convolutional} & 2 x CNN (2 x 2) + 2 x LSTM (128) + 2 x FCN (512) & $595$ & $.329\pm.02$\\
        &       & 2 x CNN (2 x 32) + 2 x LSTM (128) + 2 x FCN (512) & $607$ & $.350\pm.03$\\
        &       & 2 x CNN (2 x 128) + 2 x LSTM (128) + 2 x FCN (512) & $611$ & $.313\pm.01$\\
        \cline{2-5}

        & \textbf{3D-CNN-DNN (proposed)}  & 2 x CNN (2 x 2 x [2, 32, 128]) + 2 x FCN (512) & $792 - 810$  & $.250$\\ 
        &       & 3 x CNN (2 x 2 x 2) + 2 x FCN (512) & $798$ & $.480\pm.03$\\
        &       & 3 x CNN (2 x 2 x 32) + 2 x FCN (512) & $793$  & $.484\pm.05$\\
        &       & 3 x CNN (2 x 2 x 128) + 2 x FCN (512) & $807$  & $\mathbf{.496\pm.03}$\\
        \cline{2-5}

        & \textbf{3D-CNN-DNN-ELM (proposed)}  & 2 x CNN (2 x 2 x [2, 32, 128]) + 2 x FCN (512) & $527 - 548$  & $.250$\\ 
        &       & 3 x CNN (2 x 2 x 2) + 2 x FCN (512) & $528$ & $.495\pm.05$\\
        &       & 3 x CNN (2 x 2 x 32) + 2 x FCN (512) & $531$  & $\mathbf{.516\pm.02}$\\
        &       & 3 x CNN (2 x 2 x 128) + 2 x FCN (512) & $546$  & $.512\pm.03$\\
\hline
\end{tabular}
\vspace{0.15cm}
\caption{UA of speaker independent experiments, CNN: Convolutional network (resolution), FCN: Fully-connected network (\#node), LSTM: Long-short-term-memory (\#cell), ELM: Extreme-learning-machine}\label{tab:comparison}
\end{table*}

\subsection{Exploring resolutions of kernels}
Table \ref{tab:resolution} summarises results of variations in temporal and spectral depth (L x T x S) of kernels. Commonly, we use three convolutional layers. In (a), we use shallow temporal kernels but vary spectral depth to find the optimal spectral depth. Shallow or moderately deep spectral kernels (2, 32, 128) significantly outperform very deep kernels (256). We assume that too deep kernels may cause over-fitting. In (b), we use shallow spectral depth but vary temporal depth. No matter which temporal depth changes, they perform worse than many cases with deep spectral depth in (a). Moreover, based on the result of (a), we keep deep spectral depth (128) but vary temporal depth. A deep long-term kernel (8 x 2 x 128) results in the best performance among them but it does not outperform the architecture with shallow temporal and deep spectral kernels (a). From these results, we empirically conclude that the kernels with shallow temporal depth and moderately deep spectral depth (2 x 2 x 128) work the best for 3D CNNs in our task. However, except for the very deep spectral kernel (256), the gap is not significant. Hence, we will examine spectral depth of 2, 32, and 128 but drop that of 256 for other methods in subsequent experiments.

\subsection{Compared to state-of-the-art}
We compare our proposed methods to state-of-the-art methods: DNN-ELM~\cite{kunHan2014dnn}, LSTM-ELM~\cite{lee2015high}, 1D-CNN-LSTM~\cite{trigeorgis2016adieu}, 2D-CNN-LSTM~\cite{anandconvoluted}, and 2D-CNN-LSTM-DNN~\cite{sainath2015convolutional}. The following descriptions explain architectures and features for the state-of-the-art methods.

\textbf{DNN-ELM~\cite{kunHan2014dnn} and LSTM-ELM~\cite{lee2015high}.} They use off-the-shelf features: F0, voice probability, zero-crossing-rate, 12-dimensional (D) MFCC with Root Mean Squared energy and those first time derivatives (totaling 32 features). Only DNN-ELM uses a contextual windows of five frames to model temporal dynamics. DNN-ELM has three hidden layers of 256 nodes, and LSTM-ELM has two hidden layers with 128 cells. Statistical functionals to extract utterance-level features and a proceeding Extreme Learning Machine (ELM) have the same setting in \cite{kunHan2014dnn,lee2015high}.

\textbf{1D-CNN-LSTM~\cite{trigeorgis2016adieu}.} We extract a 32000D vector at every 2s long sequence and segment each vector to 20 sub-sequences using a contextual window with 40ms (1600D). A first temporal convolutional layer has a length of 80, followed by a max-pooling layer with a size of 2. Next, a second temporal convolutional layer with a length of 800 followed by a max-pooling with a size of 40. LSTM blocks with two hidden layers are stacked on the top conversational layer, and the cell size varies (128 and 256).

\textbf{2D-CNN-LSTM~\cite{anandconvoluted} and 2D-CNN-LSTM-DNN~\cite{sainath2015convolutional}.} The shape of feature vectors is equal to that of the proposed method (10 x 10 x 256). As the same way of our proposed method, the homogeneous kernels are adopted, and we only vary the spectral depth (2, 32, and 128). A max-pooling layer follows each convolutional layer. The first pooling layer has a resolution of 2 x 2 and the second one has that of 4 x 4. Two LSTM layers that have a cell size of 128 are stacked on the top conversational layer. Lastly, two fully-connected layers with 512 nodes are stacked too.

Table \ref{tab:comparison} summarises results. For comparison, it includes the performance of 3D-CNN-DNNs using the kernels of 2 x 2 x 2, 2 x 2 x 32, and 2 x 2 x 128 (previously presented in Table \ref{tab:resolution}), too.  2 x 2 x [2, 32, 128] is short for these three shapes of kernels. 

Any configuration of 2D-CNN-LSTM-DNN and 2D-CNN-LSTM does not outperform DNN-ELM and LSTM-ELM. While emotional classes can be directly learned from spectrogram features~\cite{ghoshrepresentation2016}, outperforming off-the-shelf features is still challenging. Moreover, later experiments show that 2D-CNN-LSTM(-DNN) could not avoid critical over-fitting problems as increasing the depth from two to three. Because of the complexity of LSTM, the depth is limited to two, and it might not be sufficiently deep to learn the complicated sequential structure of emotional speech ~\cite{kim2013emotion}. 1D-CNN-LSTM using raw waveform does not outperform LSTM-ELM, too. We could not improve the performance by increasing the depth from two to three. We assume that it is mainly due to the huge number of parameters. We also differentiate other configuration (e.g. resolutions of pooling layers) later but any further gain is not observed. 

On the other hand, 3D-CNN-DNN and 3D-CNN-DNN-ELM, outperform DNN-ELM and LSTM-ELM with significant gains ($\mathbf{11 \sim 13}$ and $\mathbf{7 \sim 9}$\%, respectively). Moreover, they outperform 1D-CNN-LSTM using raw waveform by $\mathbf{10 \sim 12}$\%. When we use the depth of two, the trained models classify all test samples as ``neutral'' regardless of its kernel resolutions, resulting in UA of $.25$. However, they show the best performance at the depth of three. With the given data set, the shallow temporal but moderately deep spectral kernels are optimal for 3D-CNN-DNN (2 x 2 x 128) and 3D-CNN-DNN-ELM (2 x 2 x 32). We could not observe any gains as increasing the depth from three. Although 3D-CNN-DNN-ELM outperform 3D-CNN-DNN, the difference is not significant ($p = .31$). 

\begin{figure}[!th]
\begin{minipage}[b]{0.48\linewidth}
  \centerline{\includegraphics[width=3.9cm,height=3.1cm]{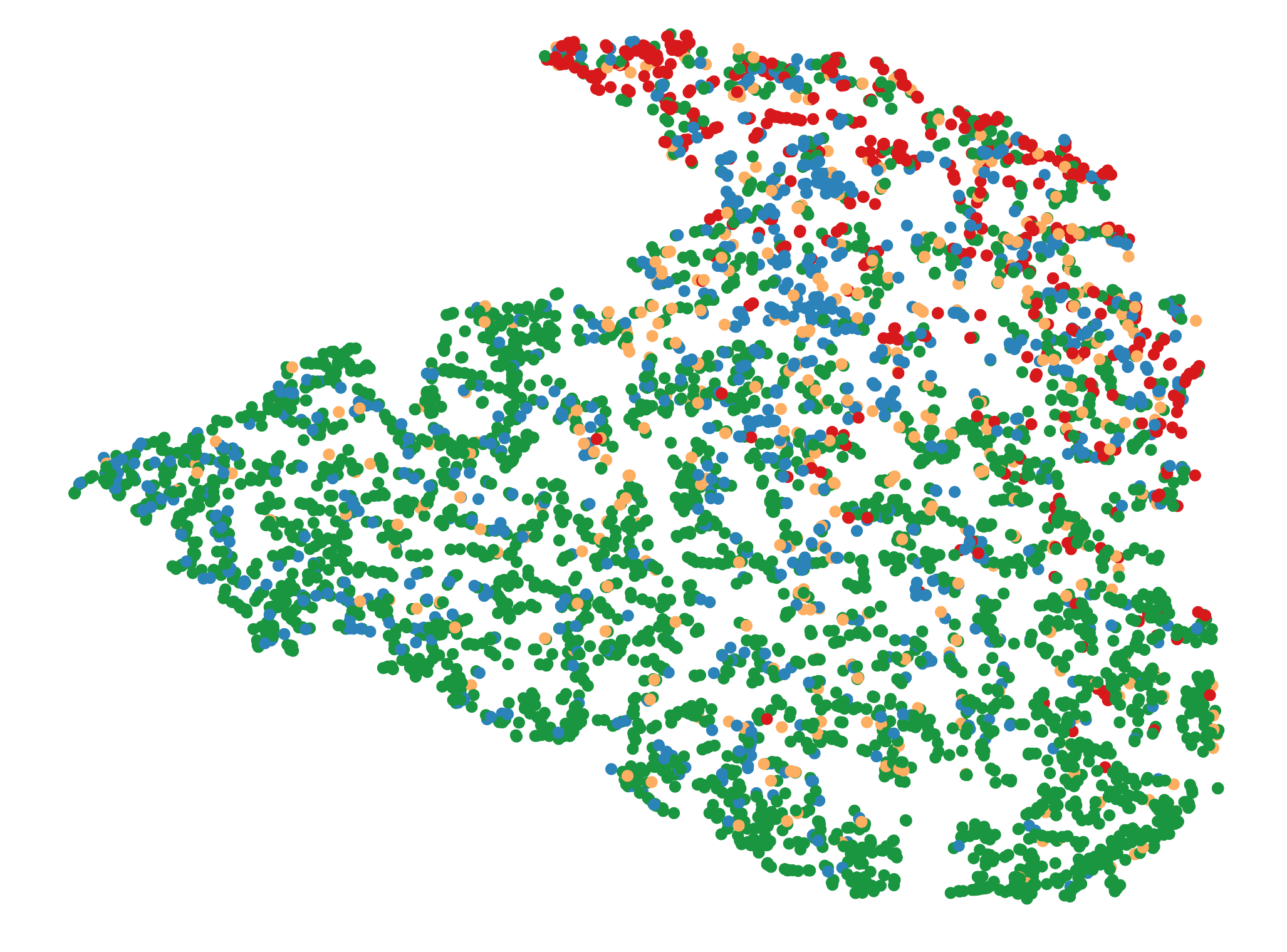}}
  \centering{(a) DNN-ELM}
\end{minipage}
\begin{minipage}[b]{0.48\linewidth}
  \centerline{\includegraphics[width=3.9cm,height=3.1cm]{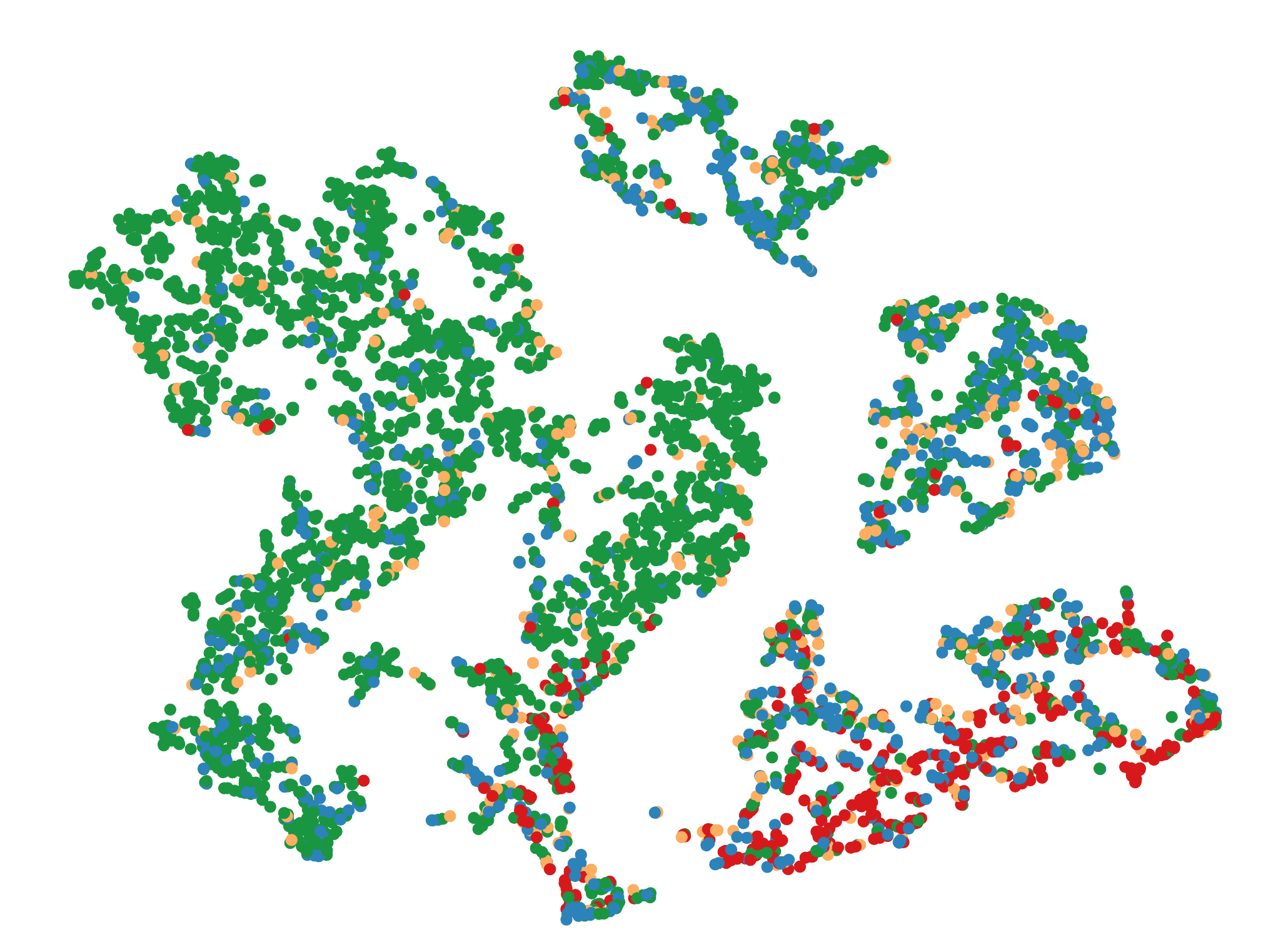}}
  \centering{(b) LSTM-ELM}
\end{minipage}
\begin{minipage}[b]{0.48\linewidth}
  \centerline{\includegraphics[width=3.9cm,height=3.1cm]{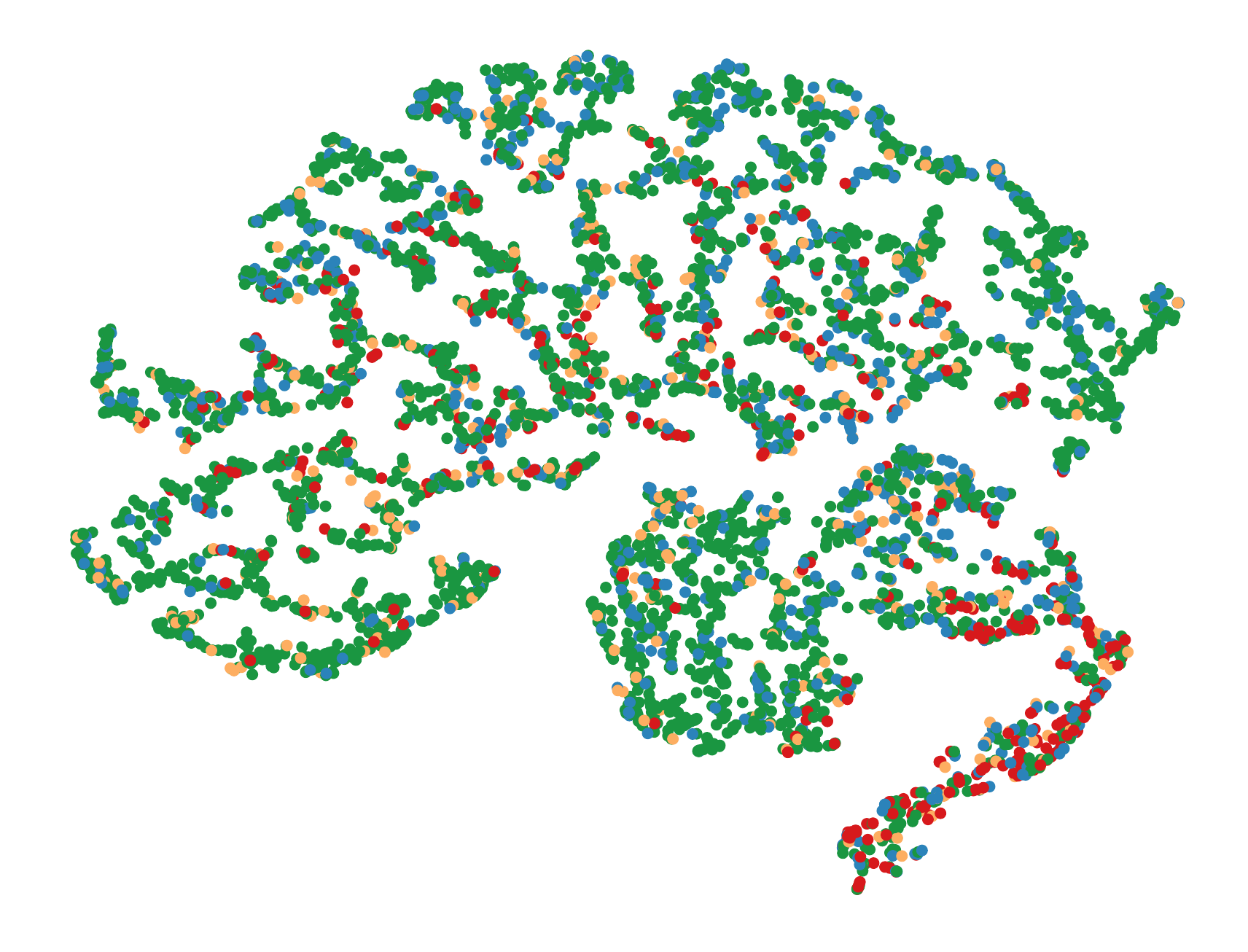}}
  \centering{(c) 2D CNN-LSTM-DNN}
\end{minipage}
\begin{minipage}[b]{0.48\linewidth}
  \centerline{\includegraphics[width=3.9cm,height=3.1cm]{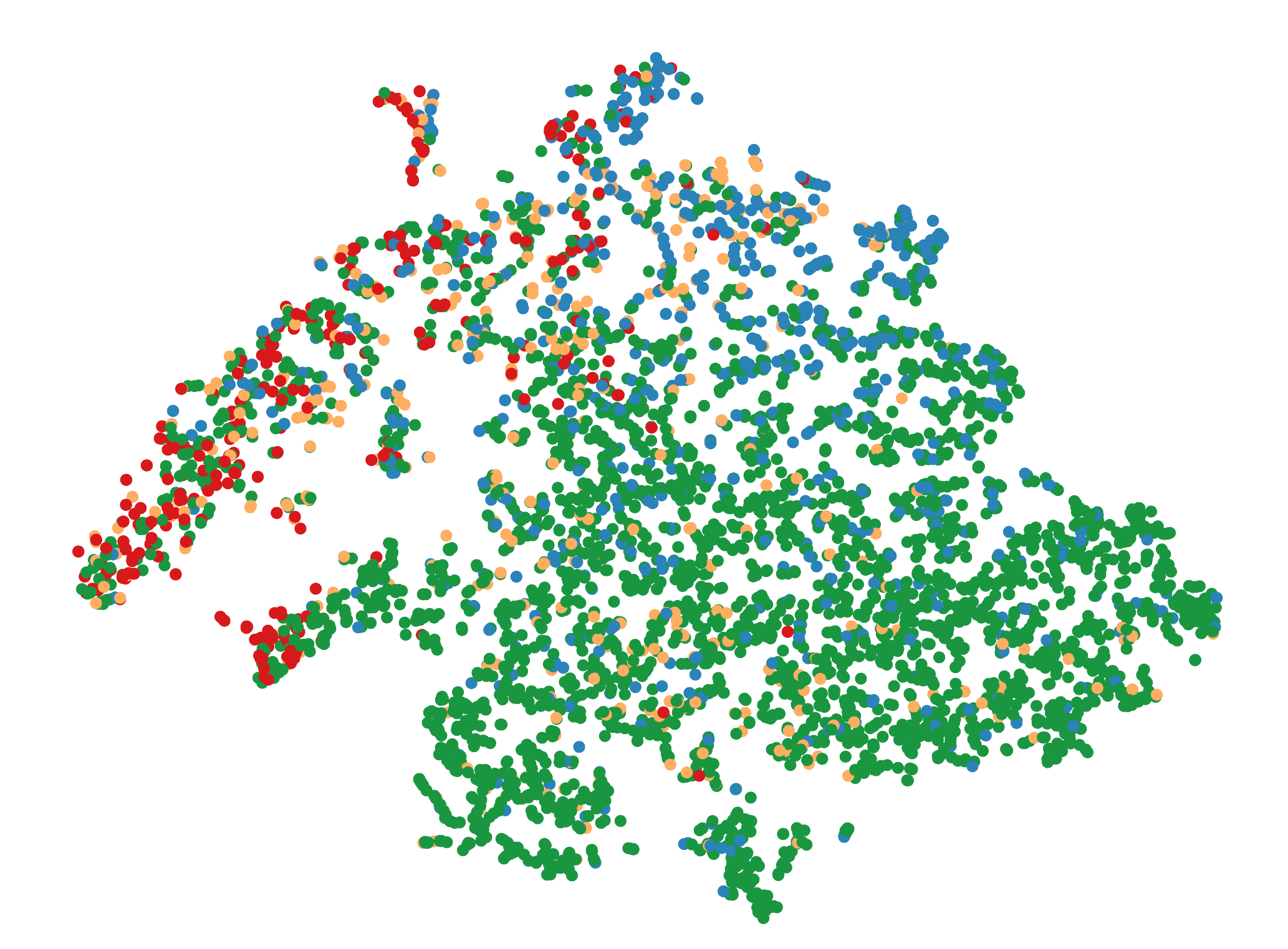}}
  \centering{(d) 3D CNN-DNN}
\end{minipage}
\caption{The result of T-SNE for the learned features of the test data; coloured by emotional categories (green: neutral, orange: happy, blue: sad, and red: angry)}
\label{fig:tsne}
\end{figure}

\begin{table}[!th]
\centering
\begin{tabular}{llllllllll}
\hline
\multicolumn{10}{c}{Method} \\
\hline
\multicolumn{5}{c}{DNN-ELM} & \multicolumn{5}{c}{LSTM-ELM}\\
\hline
  & N & H & S & A    &   & N  & H & S & A\\
\hline
N & 93 &  0 & 4 & 3     & N & 83 &  4 & 4 & 9 \\
H & 82 &  0 & 7 & 11    & H & 59 &  10 &12 &  19 \\
S & 26 &  0 & 71 & 3    & S & 23 &  4 & 69 &  4 \\
A & 80 &  0 & 11 & 9    & A & 51 &  4 & 14 &  32 \\
\hline
\hline
\multicolumn{5}{c}{2D-CNN-LSTM-DNN} & \multicolumn{5}{c}{\textbf{3D-CNN-DNN}}\\
\hline
  & N & H & S & A    &   & N  & H & S & A\\ 
\hline
N & 80 & 2 & 10 & 8  & N & \textbf{88} &  0 & 7 & 4 \\
H & 72 & 1 & 12 & 15 & H & 64 & \textbf{8} &  8 & 20 \\
S & 38 & 0 & 58 & 4  & S & 18 & 1 & \textbf{75} & 6 \\
A & 65 & 0 & 13 & 22 & A & 41 & 1 & 8 & \textbf{50} \\
\hline
\end{tabular}
\vspace{0.15cm}
\caption{Confusion matrix (\%). N: Neutral, H: Happy, S: Sad, A: Angry.}\label{tab:cm}
\end{table}

Next, we investigate how each class becomes discriminative in the feature space. To this end, we visualise utterance-level representations of one test set of our cross-validation by using t-distributed stochastic neighbour embedding (T-SNE)~\cite{maaten2008visualizing}. T-SNE is a non-linear transform technique to embed high-dimensional data into a space of two or three dimensions. We obtain utterance-level features from the top fully-connected layers (before the softmax layer). Figure \ref{fig:tsne} shows results of DNN-ELM, LSTM-ELM, 2D-CNN-CNN-LSTM and 3D-CNN-DNN, and Table \ref{tab:cm} presents their confusion matrices. Compared to 2D-CNN-LSTM-DNN, 3D-CNN-DNN shows a more discriminative separation between ``neutral'' and ``anger''. In Table \ref{tab:cm}, confusion between ``neutral'' and ``anger'' significantly decreases in 3D-CNN-DNN. The gains are $\mathbf{8}$\ and $\mathbf{28}$\%, respectively. Classification of ``neutral'' and ``sadness'' shows the similar result too. The gain for ``sadness'' is $\mathbf{17}$\%.

\subsection{Discussion}
Our data has a large set of speakers from multiple corpora that have different conditions of recordings, tasks, and \etc. Moreover, the data has an imbalanced distribution of the categories and our evaluation is carried out in a speaker-independent manner. These settings are close to realistic challenges and pose a great potential of over-fitting (caused by the huge variance). In such a harsh condition, LSTM modelling temporal dynamics of emotional speech seems inefficient. Indeed, 1D-CNN-LSTM with a large number of parameters is vulnerable to over-fitting. Emotional speech corpora in the research community are inevitably limited compared to other tasks (speech and image recognition). Compared to the complex augmented architectures, 3D CNNs have relatively simpler architectures, and those with a moderate number of parameters could learn spectro-temporal features in a seamless way. Lastly, the variance of the performance shows the importance of large-scale cross-validation and statistical tests. While they should not be neglected, it is arduous to optimise deep architectures with the great variance of the number of samples and that of class distribution from aggregated corpora. We believe that our evaluation is able to present the realistic performance in the wild.


\section{Conclusions and future work}\label{sec:conclusion}
In this paper, we proposed deep 3-dimensional convolutional networks (3D CNNs) based methods to learn spectro-temporal features for the task of speech emotion recognition (SER). We designed 3D CNNs to learn short and long-term spectro-temporal features with a moderate number of parameters. We evaluated the proposed and other state-of-the-art methods using large-scale speaker independent experiments. First, we found that shallow temporal and moderately deep spectral kernels are optimal to the explored homogeneous architectures. Next, we found that 3D CNNs are more suitable for spectro-temporal feature learning compared to other CNN based methods (e.g. CNN-LSTM). CNN-LSTMs using low-level representations do not outperform methods using off-the-shelf features. However, 3D CNNs learn the features with a moderate number of parameters and significantly outperform the other methods. In addition, we visualised the spectro-temporal features learned by the method via T-distributed stochastic neighbor embedding technique. More discriminative clusters of emotional classes can be observed in the feature space, and our proposed method significantly decreases the confusion rates. As future work, we plan to investigate identity skip-connections that are recently popular for optimising deep architectures \cite{veit2016residual}.

\ifCLASSOPTIONcompsoc
  \section*{Acknowledgments}
\else
  \section*{Acknowledgment}
\fi

The research leading to the results was supported by the European Community's 7th Framework Programme under Grant agreement 610532 (SQUIRREL - Clearing Clutter Bit by Bit) and 611153 (TERESA - Telepresence Reinforcement-learning Social Agent).
\clearpage
\small
\bibliographystyle{IEEEtran}
\bibliography{mybib}

\end{document}